\documentclass{article}
\usepackage{spconf,amsmath,graphicx}
\usepackage{hyperref}

\usepackage{amsmath}
\usepackage{amssymb}
\usepackage{graphicx}
\usepackage{subfigure}
\usepackage{multirow}
\usepackage{booktabs}
\usepackage{bbm}

\title{Towards Spatio-temporal Sea Surface Temperature Forecasting via Static and
Dynamic Learnable Personalized Graph Convolution Network}
\name{Xiaohan Li$^{1}$ \qquad Gaowei Zhang$^{2}$ \qquad Kai Huang $^{1}$ \qquad Zhaofeng He$^{2}$ \thanks{Zhaofeng He Xu is the corresponding author.}}
\address{ $^1$School of Science, Beijing University of Posts and Telecommunications, Beijing\\
        $^2$School of Artificial Intelligence, Beijing University of Posts and Telecommunications, Beijing}

\usepackage{booktabs}
\usepackage{multirow}

\begin{document}
%
\maketitle

\begin{abstract}
    Sea surface temperature (SST) is uniquely important to the Earth's atmosphere since its dynamics are a major force in shaping local and global climate and profoundly affect our ecosystems. Accurate forecasting of SST brings significant economic and social implications, for example, better preparation for extreme weather such as severe droughts or tropical cyclones months ahead. However, such a task faces unique challenges due to the intrinsic complexity and uncertainty of ocean systems. Recently, deep learning techniques, such as graphical neural networks (GNN), have been applied to address this task. Even though these methods have some success, they frequently have serious drawbacks when it comes to investigating dynamic spatiotemporal dependencies between signals. To solve this  problem, this paper proposes a novel static and dynamic learnable personalized graph convolution network (SD-LPGC). Specifically, two graph learning layers are first constructed to respectively model the stable long-term and short-term evolutionary patterns hidden in the multivariate SST signals. Then, a learnable personalized convolution layer is designed to fuse this information. Our experiments on real SST datasets demonstrate the state-of-the-art performances of the proposed approach on the forecasting task. 
\end{abstract}

\section{Introduction}

As a key physical attribute of the world's oceans, sea surface temperature (SST) influences almost every aspect of the Earth's atmosphere \cite{alexander2018projected}. SST changes can profoundly affect local and global climate \cite{annamalai2005impact}. For instance, SST changes, particularly those abnormal changes, would inevitably cause dramatic fluctuations of the atmospheric water vapor levels, which produce various precipitation patterns leading to extreme weather such as heavy rain, severe drought, and tropical cyclones \cite{hoerling2003perfect}. These extreme weathers often result in serious socioeconomic impacts such as power system outages, property damages, and life losses \cite{meehl2000introduction}. Moreover, SST changes also significantly impact the biosphere, particularly the plant, animals, and microbes in marine ecosystems. For example, researchers have found that positive SST anomalies are widely associated with mass coral mortality events in the northeastern Caribbean Sea \cite{winter1998sea}, and many other places around the world.

Given SST dynamics' strong correlations with many natural phenomena which have far-reaching impacts on our social and economic systems, accurate SST forecasting could help government and environment agencies plan months ahead for many scenarios, such as precipitation monitoring, marine life protection, tourism, fishery, and so on \cite{meehl2021initialized}. However, forecasting SST is non-trivial. The global ocean system is a typical complex open giant system; thus, numerous factors influence SST, e.g., the absorption of energy in Sunlight, human activities, local geological structure, and so on \cite{o2019observational}. Besides, global warming makes SST anomalies appear much more frequently than before, introducing new challenges to SST forecasting \cite{zha2022multiple}.

The development of SST forecasting techniques has gone through three phases. Initially, researchers used statistical time series models to accomplish this task \cite{li2012new}. Later, machine learning techniques were introduced into the field \cite{lins2013prediction}. And recently, deep learning techniques have become the preferred choice for many existing works due to their flexibility and performance. Almost all important families of deep learning techniques, e.g., recurrent neural networks (RNN) \cite{xie2019adaptive}, long short-term memory (LSTM) \cite{kim2020sea}, have been applied to SST forecasting. However, due to the limitation of their network structures, RNNs and LSTMs cannot naturally capture and express the complex spatio-temporal dependencies among SST data, especially those involving irregular spatial data \cite{yuan2020deep}. Recently, the graph neural network (GNN) model extends the neural network approach to handle spatial dependencies between variables \cite{sun2021time}.\cite{yu2020superposition} proposed SGNN (superposition graph neural network) to make full use of spatio-temporal features for prediction through the composition of geographic location and other relevant information. However, the above graph neural networks all assume that a well-defined graph structure is already in place. The lack of flexibility in the graph structure further limits the generality and realistic applicability of the above GNN-based models due to the poor applicability in dealing with SST data with anomalous behavior. Meanwhile, a time series is influenced by both the input series and its own evolutionary law \cite{qin2017dual}, and the current spatiotemporal GNNS all ignore the evolutionary law of the variables themselves.


To address these issues, we propose a static and dynamic learnable personalized graph convolution network (SD-LPGC) to capture  dynamic spatio-temporal dependencies in SST data. Specifically, as shown in Figure 1, we simultaneously infer two different networks represented by graph adjacency matrices. The first one is a trainable node-embedding network (static network), used to model long-term patterns between variables. The second one is a dynamic graph network, used to capture the changing short-term patterns in SST data. In the learning process of dynamic graphs, we employ a gating mechanism to integrate input sequences and node embeddings and use the learned static graphs as a bias for generalization.  In the learnable personalized graph convolution module, we model the input sequences separately in combination with static and dynamic graph structures in order to efficiently integrate the input sequence information with the nodes' own evolutionary patterns.  

In summary, our main contributions are as follows: 1) A dynamic graph learning network, which effectively captures spatial-temporal dependencies in geo-coded, multivariate time series data and learns long-term patterns and constantly-changing short-term patterns for accurate prediction in multi-graph structure learning; 2) A learnable personalized graph convolution module that models the individual evolutionary patterns and the influence of its neighboring nodes in the graph, and designs self-restart probabilities to model the dynamics and heterogeneity of fusion patterns; 3) A novel end-to-end SST forecasting framework (SD-LPGC), which synthesizes static, dynamic graph learning, temporal convolution, and dynamic personalized graph convolution to deliver promising forecasting for multiple-domain, geo-coded, multivariate time series data, e.g., SST data; 4) Experiments with real-world SST data, which demonstrate SD-LPGC's state-of-the-art performance in SST forecasting and identify human-understandable long-term and short-term patterns; 5) Two geo-coded, multivariate SST datasets, which has been carefully curated and made publicly available at: \url{https://figshare.com/articles/dataset/SST_rar/20496285}.
\begin{figure}[t]
\centering
\includegraphics[width=\columnwidth]{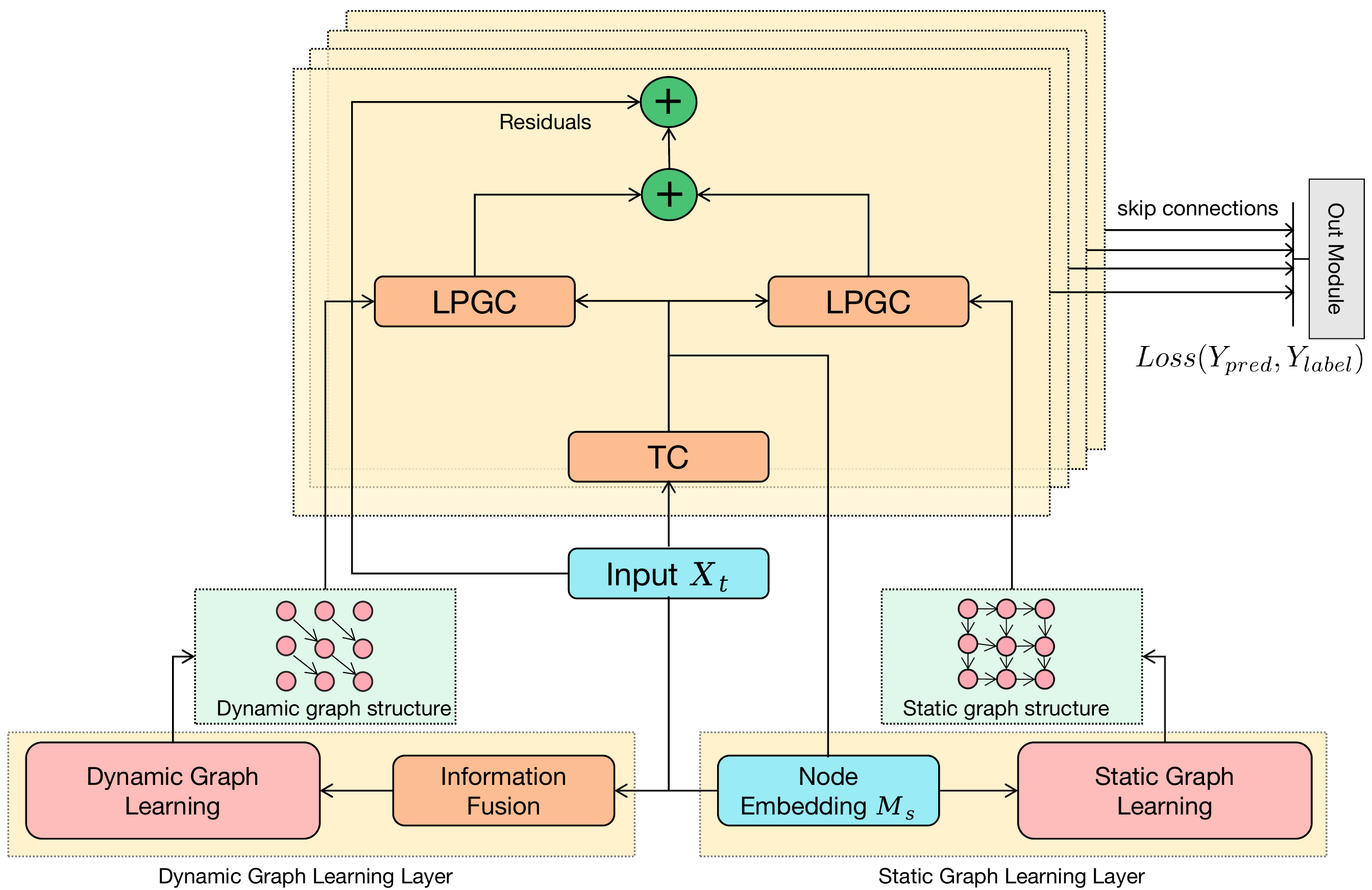}                 
\caption{Overview of the proposed SD-LPGC. It consists of (1) two graph learning layers, (2) $K$ temporal convolution modules(TC), (3) $K$ learnable personalized graph convolution modules(LPGC), and (4) an output module. The input data is first feed into the graph learning layers to generate the static and dynamic adjacency matrix. Then all the input data, the learned node embeddings, and the generated adjacency matrixes go through the $K$ layers TC Module and LPGC Module for feature transformation. Finally, the output module is employed to produce the SST predictions.}\label{fig1}
\end{figure}

\section{Methodology: SD-LPGC}

\subsection{Overview of the Architecture of SD-LPGC }
We first introduce the general framework of SD-LPGC's architecture. As shown in Figure \ref{fig1}, the SD-LPGC consists of four parts:(1) two graph learning layers, (2) $K$ temporal convolution modules (TC), (3) $K$ learnable personalized graph convolution modules (LPGC), and (4) an output module. The graph learning layers can generate two different types of adjacency matrices, namely, static adjacency matrix and dynamic adjacency matrix. They are responsible for capturing long-term and short-term spatial dependency weights between SST observation stations, respectively. The temporal convolution module adopts the dilated inception layer as in previous work \cite{wu2020connecting} to extract temporal dependencies. The learnable personalized graph convolution module performs information aggregation based on the adjacency matrices generated by the graph learning layer, respectively. We are going to introduce (1) and (3) in detail. Since (2) is a previous work and (4) is a simple linear network, we do not present these two parts in order to keep the paper concise.

\subsection{Graph Learning Layer}
Inferring the time-aware graph structure is challenging for the following reasons: the spatial-temporal dependencies between variables are time-varying, and these dependencies are the result of heterogeneous information synthesis and fluctuate up and down around long-term patterns.

To address these issues, we first design a static graph learning layer that uses learnable node embeddings $M^s\in R^{N\times d}$ to infer the long-term patterns of the graph $A^s=M^s\cdot M^{s\top}$, where $A^s\in R^{N\times N}$. Then, we use Relu and Softmax functions to remove negative values and normalize them to obtain the static graph structure $\hat{A}^s$, as shown in Eq. (\ref{as}). 
\begin{equation}
\hat{A}^s=\operatorname{SoftMax}(\operatorname{ReLU}(A^s))
\label{as}
\end{equation}
Second, we design a dynamic graph learning layer to infer short-term patterns of graphs, model heterogeneous information as static structural information and dynamic semantic information, based on dynamic input data to infer the dynamic graph matrix and use the learned long-term patterns as the a priori bias of the dynamic graph matrix. In the dynamic graph learning layer, we first use 1 × 1 standard convolution to extract the features as dynamic semantic information, converting the input to the same dimension as $M^s$, i.e., $X_T = \operatorname{Conv}(X_{t-H+1:t})$. To better fuse the static structural information $M^s$ with the dynamic semantic information $X_T$, similar to previous works\cite{LI2023119374} we used GRU to obtain the fused node representation $h_r$ within the input time window H. Then, we used the inner product operation to infer the edge weights between nodes and proposed a multi-headed version to stabilize the learning process. As shown in Eq. (\ref{dropout}).

\begin{equation}
E^{k}_{v^i,v^j}=\operatorname{Dropout}{(\frac{<f_1(\mathrm{LN}(h_r^i),f_2(\mathrm{LN}(h_r^j))>}{\sqrt{d_k}})}
\label{dropout}
\end{equation}

Where LN is LayerNorm, $f_1$, $f_2$ represent two different nonlinear projections in the $k^{th}$ head, and $d_k$ is the dimensionality of each head.$<{\cdot},{\cdot}>$ denotes the inner product operator and $E^k$ denotes the edge weight of the $k^{th}$ subspace between vertices $v^i$ and $v^j$. Here we include Dropout to force the model to learn a more general pattern of variation. Finally, we take the $K$ spatial similarity results as the final edge weights, as shown in Eq. (\ref{add}).
\begin{equation}
\hat{E}_{v^i,v^j}=\sum_{k=1}^{K}{E^{k}_{v^i,v^j}}+\mathrm{LN}(f_3(h_r^i),f_4(h_r^j))
\label{add}
\end{equation}
where $f_3$ denotes the shared trainable weights. The first term of the equation is used to obtain the association and the second term is used as a skip connection to make the model easier to learn.

In matrix-level fusion, we use long-term patterns inferred from the static graph learning layer as a priori bias and use LayerNorm and Dropout to train faster and prevent overfitting problems.Finally, we use RelU to eliminate negative values in the correlation and use SoftMax for normalization to obtain the dynamic graph structure $\hat{A}^d$, as shown in Eq. (\ref{ad}).
\begin{equation}
\hat{A}^d=\mathrm{SoftMax}(\mathrm{ReLU}(\operatorname{Dropout}(\mathrm{LN}(\hat{E}))+\hat{A}^s))
\label{ad}
\end{equation}
where $\hat{A}^d_{v^i,v^j}$ denotes the final inferred relationship between nodes $v^i$ and $v^j$ within the input time window H.

\subsection{Learnable Personalized Graph Convolution Module}

In this paper, we consider multivariate time series as a nonlinear autoregressive exogenous model, in which the variables are not only influenced by the information of the input series but also subject to their own evolutionary laws. The current spatiotemporal GNNS has ignored the self-evolution pattern of variables. Therefore, we design the learnable personalized graph convolution module (LPGC module) to obtain information from static and dynamic graphs, respectively, in order to effectively model the interaction of long-term and short-term patterns.

Given dynamic and static graph adjacency matrices, the backbone process of the LPGC module can be expressed as follows:

\begin{equation}
\begin{gathered}
Z^0 = f_{\theta} (\hat{X})\\
Z^{l+1}=\left(1-\alpha^{ l}\right) A^* Z^k+\alpha^{l} \ddot{H} 
\label{yyy}
\end{gathered}
\end{equation}
where  $\hat{X}$ is the output of the TC module, the TC module in this paper adopts  the same structure as the previous work\cite{wu2020connecting},$\alpha^{l}$ denotes the influence of the nodes themselves, $L$ defines the number of power iteration steps, and $l \in [0, L-1]$. $A^*$ represents the static adjacency matrix $A^s$ and the dynamic adjacency matrix $A^d$, where $A^s$ is from Eq. (\ref{as}), $A^d$ is from Eq. (\ref{ad}). $\ddot{H}$ denotes the evolutionary pattern of the node itself.  Eq. (\ref{yyy}) measures the influence of the nodes themselves and their neighbors in each information dissemination step.

\begin{table}[b]
\centering
\resizebox{0.48\textwidth}{!}{
\begin{tabular}{lrrrrr}
\hline
Dataset & \#Samples & \#Nodes & Sample Rate & Input Length $u$ & Output Length $v$ \\ \hline \hline
Bohai Bay SST &  2189& 136 & 1day & 12 & 12\\
South China Sea SST  & 2556 & 461 & 1day  & 12 & 12\\

     \hline
\end{tabular}}
\caption{Dataset statistics.}
\label{datasets}
\end{table}

\begin{table*}[htbp]
\resizebox{1.0\textwidth}{!}{
\begin{tabular}{llrrrrrrrrrrrr}
\hline
\multirow{2}{*} {Model}      & Horizon         & \multicolumn{3}{c}{3} & \multicolumn{3}{c}{6} & \multicolumn{3}{c}{9} & \multicolumn{3}{c}{12} \\ \cline{2-14} 
              & Metrics         & MAE   & RMSE  & MAPE(\%)  & MAE   & RMSE  & MAPE(\%)   & MAE   & RMSE  & MAPE(\%)   & MAE   & RMSE   & MAPE(\%)   \\ \hline
\multicolumn{2}{l}{Seq2Seq}     &   2.34  & 2.79   &   55.23         &   2.56  & 3.11    &  68.24          &     2.84  &    3.47& 73.26        &     3.08  &  3.76 & 84.28         \\ \hline
\multicolumn{2}{l}{GAT-Seq2Seq} &    2.21   &  2.65     &   47.22    & 2.43      &     2.85  & 63.23      &  2.69     &    3.25   &  69.25    & 2.89      & 3.51       &    78.27   \\ \hline
\multicolumn{2}{l}{DCRNN}       &   2.51    &  3.10     & 68.77   &   2.65    &   3.24     &  72.73 &     2.83  &   3.43    &   77.98  &    3.03   &  3.65      &    90.78 \\ \hline
\multicolumn{2}{l}{ST-MetaNet}  &   2.09    &  2.52       &  41.77    & 2.20      & 2.69     & 50.22 &    2.40   &    2.96   &    59.23   &    2.62       &      3.24  &   71.25   \\ \hline

\multicolumn{2}{l}{MTGNN}    &  1.79  &  2.10    &  50.32  &  2.06    &    2.41  &   53.41  &        2.38   & 2.79 &   57.30 &    2.58  &   2.96  &    72.95     \\ \hline

\multicolumn{2}{l}{GTS}    &  1.70  &    2.12   &    43.50  &    1.93   &    2.33  &     47.90  &    2.23   &   2.85    &    49.32   &   2.62   &   3.06  &  79.81          \\ \hline
\multicolumn{2}{l}{DGCRN}       &  1.54    &  1.96    &  33.49  &   1.59    &    2.02   &    34.14 &    1.64 &   2.08   &  34.82   &   1.69    &    2.15   &   35.64     \\ \hline
\multicolumn{2}{l}{ST-Norm}     &   0.79    &    1.11  &  29.52   &  0.88    &    1.20   &   34.77  &  0.94     &   1.29    &   37.99 &    1.02   &    1.33     &  40.90 \\ \hline
\multicolumn{2}{l}{SD-LPGC}      &   $\mathbf{0.55 }$ &  $ \mathbf{ 0.72 } $ & $\mathbf{  14.47 }$   &$\mathbf{ 0.67} $    &$\mathbf{0.88 } $& $\mathbf{17.23 }$ &  $  \mathbf{ 0.74 }$ & $  \mathbf{  0.98}$  &  $\mathbf{17.56 }$ & $ \mathbf{ 0.83  } $& $\mathbf{  1.08 }  $  &   $\mathbf{24.43}$  \\ \hline
\end{tabular}}
\caption{Baseline comparison on Bohai bay SST dataset.}
\label{bohai}
\end{table*}

We first introduce the self-evolutionary pattern of nodes $\ddot{H}$. The module consists of an embedding layer and multiple MLP modules. For simplicity, we refer to a fully connected layer by the abbreviation FC(·). We first transform the input using a fully-connected layer, i.e., $H= \mathrm{FC}_{1}(X)$. Then, by using initial 
node embedding matrices $M^s$, we attach $\hat{H}= \mathrm{Concat}(H, M^s)$ denotes the hidden representation with temporal and spatial identities. Then, we employ a MLP module to extract features,i.e.
$\ddot{H}=\mathrm{FC}_{2}(\mathrm{FC}_{3}\left(\operatorname{ReLU}\left(\mathrm{FC}_{4}\left(\hat{H}\right)\right)\right)+\hat{H})$. Here, residual connections are added to each layer to make the model easier to train. To model the dynamics and heterogeneity of the integration pattern of its own evolutionary pattern $\ddot{H}$ and the input sequence affecting $Z^l$, we designed the self-restart probability $\alpha^l$, which is computed by another fully connected layer,i.e.$\alpha^{l} = \mathrm{Sigmoid}(\mathrm{FC}_{5}(\ddot{H} + Z^l))$.A larger $\alpha^{l}$ means that the self-evolutionary pattern of node $i^{th}$ is more important. The determination of the weights is based on the self-evolutionary pattern and the characteristics of the aggregation.

The output is a sequence of hidden states $[{Z}^{0}, {Z}^{1}, ..., {Z}^{(l-1)}]$, which captures the external information based on the input sequence and the evolutionary pattern of the nodes themselves from low to high. Combine with Eq. (\ref{yyy}), the final learnable personalized graph convolution is as Eq. (\ref{final_dg}):

\begin{equation}
Z= \mathrm{FC}_{6}(\mathrm{Concat}(Z^0, Z^1, ..., Z^{(L-1)}))
\label{final_dg}
\end{equation}

To model the interaction of short-term and long-term patterns, we adopt two learnable personalized convolution modules to capture information on static and dynamic graphs separately, that is, replacing $A^*$ with learned $A^s$ and $A^d$. The final output is the sum of the two personalized graph convolutional modules.

\section{Experiments}
\subsection{Overview of the Experiments and Datasets}
We first evaluate the performances of SD-LPGC on two in-house collected \textbf{Bohai bay and South China Sea SST datasets}. Bohai bay and South China Sea are rich in hydrocarbon deposits and fisheries and have several active offshore oil fields. These all make the SST data exhibit drastic changes, thus serving  good benchmarks for evaluating SD-LPGC in dealing with complicated SST forecasting challenges. The data collection took the first author ten months, involving numerous manual cleaning and cross-checking procedures, and multiple costly field trips. 

Because it is resource and time-consuming to collect geo-coded SST data, and there are rare publicly available such datasets, we only use two dataset in the main experiment. However, in the traffic-flow forecasting domain, multiple geo-coded time-series datasets are easy to obtain. 

Among all datasets, we apply Z-Score normalization and 70\% of data is used for training, 20\% is used for testing, and the remaining 10\% for validation. In Table \ref{datasets}, we summarize statistics of benchmark datasets.

\begin{table*}[]
\resizebox{1.0\textwidth}{!}{
\begin{tabular}{llrrrrrrrrrrrr}
\hline
\multirow{2}{*} {Model}         & Horizon         & \multicolumn{3}{c}{3} & \multicolumn{3}{c}{6} & \multicolumn{3}{c}{9} & \multicolumn{3}{c}{12} \\ \cline{2-14} 
              & Metrics         & MAE   & RMSE  & MAPE(\%)  & MAE   & RMSE  & MAPE(\%)  & MAE   & RMSE  & MAPE(\%)  & MAE   & RMSE   & MAPE(\%)  \\ \hline
\multicolumn{2}{l}{Seq2Seq}      & 0.65  &   0.81  &  2.36  &   0.69    & 0.88 &   2.53 &   0.74   &    0.94  & 2.77 &  0.79  &   1.01  &  2.95 \\ \hline
\multicolumn{2}{l}{GAT-Seq2Seq}    & 0.59 & 0.76    & 2.19   &   0.65  & 0.81 &   2.24 &  0.70   &  0.88   & 2.49 & 0.73  &  0.93   &  2.83 \\ \hline
\multicolumn{2}{l}{DCRNN}       &  0.75 &  0.83   & 2.66 &      0.81  & 0.92   &  2.97 & 0.87 &  1.03   &  3.14 & 0.94 &    1.16  & 3.36  \\ \hline
\multicolumn{2}{l}{ST-MetaNet}    & 0.51  &  0.67   &  1.86  &   0.60  & 0.74 & 2.08   & 0.65    &  0.79   & 2.31 & 0.69  &  0.87   & 2.54 \\ \hline

\multicolumn{2}{l}{MTGNN}     & 0.40 &   0.52  &  1.43  &   0.48  &  0.62 &  1.72  &  0.54   &   0.70   & 1.94 &  0.59 & 0.77   &  2.12  \\ \hline

\multicolumn{2}{l}{GTS}      & 0.39 &  0.51  & 1.39   &  0.47 & 0.61 & 1.68 &   0.51   &  0.66   & 1.91 & 0.56  & 0.73 & 2.00   \\ \hline
\multicolumn{2}{l}{DGCRN}          & 0.55 &  0.72   & 1.99    & 0.61    & 0.79  & 2.19  &  0.65  &  0.84  & 2.35 & 0.69  & 0.88    &  2.47   \\ \hline
\multicolumn{2}{l}{ST-Norm}     & 0.36 &   0.47   &  1.30  &    0.49 & 0.62  &  1.76   &  0.56    & 0.71   &  2.02 &  0.63 &   0.81  & 2.27  \\ \hline
\multicolumn{2}{l}{SD-LPGC}      & $\mathbf{0.34}$ &  $\mathbf{0.44} $  &  $\mathbf{1.21} $ &   $\mathbf{0.43} $ &$\mathbf{0.56}$  &  $\mathbf{1.57} $ & $\mathbf{0.49} $  &  $\mathbf{0.63} $ &$\mathbf{1.75} $ & $\mathbf{0.52} $& $\mathbf{0.68}$  &  $\mathbf{1.88}$ \\ \hline
\end{tabular}}
\caption{Baseline comparison on South China Sea SST dataset.}
\label{nanhai}
\end{table*}

\subsection{Experimental Settings}
To verify the effectiveness of SD-LPGC, we compare it with the following representative baselines: 1) Seq2Seq \cite{sutskever2014sequence}: An end-to-end approach to sequence learning using LSTMs as encoders and decoders. 2) GAT-Seq2Seq \cite{velivckovic2017graph}: A graph attention network that utilizes masked self-attention layers to model spatial and temporal correlations with Seq2Seq, respectively. 3) DCRNN \cite{li2018diffusion}: A diffusion convolution recurrent neural network that combines diffusion graph convolution with a recurrent neural network and uses a bivariate random walk of the graph to capture spatial dependencies. 4) ST-MetaNet \cite{pan2019urban}:A deep meta-learning-based model using a meta-graph attention network and a meta-recursive neural network to capture different spatial and temporal correlations. 5) MTGNN \cite{wu2020connecting}: A model that uses graph neural networks and a designed adaptive adjacency matrix to model correlations between multivariate time series data. 6) GTS \cite{shang2021discrete}: A model that learns probabilistic graphs by optimizing the average performance of the graph distribution when the graph is unknown. 7) DGCRN \cite{li2021dynamic}: A dynamic graph convolutional recurrent network that uses and extracts dynamic features of node attributes and models the exemplary topology of the dynamic graph at each time step. 8) ST-Norm \cite{deng2021st}: A new method for separating complex external influences on multivariate time series data uses two modules, temporal normalization, and spatial normalization, to refine the high-frequency components and local components under the original data, respectively.
We compare the results of each baseline using three measures that are commonly utilized in multivariate time series forecasting. They are the Mean Absolute Error (MAE), the Root Mean Squared Error (RMSE), and the Mean Absolute Percentage Error (MAPE).
\subsection{Experimental Results}
Table \ref{bohai} and Table \ref{nanhai} shows the comparison of different models for 3 days, 6 days, 9 days, and 12 days ahead forecasting on Bohai SST dataset and South China Sea SST dataset respectively. From them, we can conclude that: (1) graph-based approaches outperform other traditional deep learning methods, indicating the sensor network is essential for SST prediction; and (2)
SD-LPGC achieves \emph{state-of-the-art} performances and its superiority is more evident in the long-term horizon (e.g., 12 days ahead). We argue that long-term SST forecasting is more beneficial in practical applications, e.g., it allows more time to take actions to prepare for potential unfavorable events associated with SST changes.
\begin{table}[htbp]
\centering
\resizebox{0.48\textwidth}{!}{
\begin{tabular}{lrrr}
\hline
Variants    & Avg-MAE    & Avg-RMSE   & Avg-MAPE(\%)   \\ \hline
SD-LPGC    & 0.66 & 0.86 & 15.91 \\ \hline
(w/o) SL   & 0.69 & 0.89 & 16.04 \\ \hline
(w/o) DL   & 0.70 & 0.91 & 16.74 \\ \hline
(w/o) LPGC & 0.72 & 0.93 & 16.96 \\ \hline
SD-GCN   &   0.71     &  0.92    &   16.92       \\ \hline
\end{tabular}}
\caption{Ablation study of SD-LPGC with different kinds of topic modeling on Bohai Bay SSL dataset. SL and DL represent the static and dynamic learning layers, respectively.}
\label{ablation}
\end{table}
\subsection{Ablation Study}
We conduct an ablation study on Bohai bay SST dataset to validate the effectiveness of key components of the developed model. The variants include 1) SD-LPGC w/o SL: SD-LPGC without Static graph learning layer. 2) SD-LPGC w/o DL: SD-LPGC without Dynamic graph learning layer. 3) SD-LPGC w/o LPGC: Only the graph learning layer and TCN are used in the developed method. 4) SD-GCN: Replace LPGC in our SD-LPGC with a normal GCN named SD-GCN.
The test results are shown in Table \ref{ablation}. Clearly, each component of our model plays an important role in SST prediction. Specifically, when we remove the graph learning layer, remove the LPGC, or replace the LPGC with a normal GCN, the prediction error suddenly increases.
\section{Conclusion}
Accurate SST forecasting will lead to enormous potential benefits to the Earth's environment and our society by opening possibilities for supporting many mission-critical tasks. This paper presents SD-LPGC, an end-to-end approach that offers state-of-the-art SST forecasting. SD-LPGC features its integration of static and dynamic graphs to learn both long-term and short-term patterns in spatio-temproal dependencies hidden in geo-coded, multivariate SST data and effectively combines SST input sequence information with the nodes' own evolutionary patterns through learnable personalized graph convolution. It reveals that correctly and adequately modeling the dependencies among variables is essential for understanding time series data. Besides, the identity of long-term and short-term patterns is highly-interpretable for human beings. Currently, we are working with the related oceanic administration agencies to deploy SD-LPGC and will continue to improve it in a real-world deployment. 

\bibliographystyle{IEEEbib}
\bibliography{strings}

\begin{thebibliography}{10}

\bibitem{alexander2018projected}
Michael~A Alexander, James~D Scott, Kevin~D Friedland, Katherine~E Mills,
  Janet~A Nye, Andrew~J Pershing, and Andrew~C Thomas,
\newblock ``Projected sea surface temperatures over the 21st century: Changes
  in the mean, variability and extremes for large marine ecosystem regions of
  northern oceans,''
\newblock {\em Elementa: Science of the Anthropocene}, vol. 6, 2018.

\bibitem{annamalai2005impact}
HSPX Annamalai, SP~Xie, JP~McCreary, and R~Murtugudde,
\newblock ``Impact of indian ocean sea surface temperature on developing el
  ni{\~n}o,''
\newblock {\em Journal of Climate}, vol. 18, no. 2, pp. 302--319, 2005.

\bibitem{hoerling2003perfect}
Martin Hoerling and Arun Kumar,
\newblock ``The perfect ocean for drought,''
\newblock {\em Science}, vol. 299, no. 5607, pp. 691--694, 2003.

\bibitem{meehl2000introduction}
Gerald~A Meehl, Thomas Karl, David~R Easterling, Stanley Changnon, Roger
  Pielke~Jr, David Changnon, Jenni Evans, Pavel~Ya Groisman, Thomas~R Knutson,
  Kenneth~E Kunkel, et~al.,
\newblock ``An introduction to trends in extreme weather and climate events:
  observations, socioeconomic impacts, terrestrial ecological impacts, and
  model projections,''
\newblock {\em Bulletin of the American Meteorological Society}, vol. 81, no.
  3, pp. 413--416, 2000.

\bibitem{winter1998sea}
A~Winter, RS~Appeldoorn, A~Bruckner, EH~Williams~Jr, and C~Goenaga,
\newblock ``Sea surface temperatures and coral reef bleaching off la parguera,
  puerto rico (northeastern caribbean sea),''
\newblock {\em Coral Reefs}, vol. 17, no. 4, pp. 377--382, 1998.

\bibitem{meehl2021initialized}
Gerald~A Meehl, Jadwiga~H Richter, Haiyan Teng, et~al.,
\newblock ``Initialized earth system prediction from subseasonal to decadal
  timescales,''
\newblock {\em Nature Reviews Earth \& Environment}, vol. 2, no. 5, pp.
  340--357, 2021.

\bibitem{o2019observational}
Anne~G O’Carroll, Edward~M Armstrong, Helen~M Beggs, Marouan Bouali,
  Kenneth~S Casey, Gary~K Corlett, Prasanjit Dash, Craig~J Donlon, Chelle~L
  Gentemann, Jacob~L H{\o}yer, et~al.,
\newblock ``Observational needs of sea surface temperature,''
\newblock {\em Frontiers in Marine Science}, vol. 6, pp. 420, 2019.

\bibitem{zha2022multiple}
Cheng Zha, Weidong Min, Qing Han, Xin Xiong, Qi~Wang, and Qian Liu,
\newblock ``Multiple granularity spatiotemporal network for sea surface
  temperature prediction,''
\newblock {\em IEEE Geoscience and Remote Sensing Letters}, vol. 19, pp. 1--5,
  2022.

\bibitem{li2012new}
Chunshien Li and Jhao-Wun Hu,
\newblock ``A new arima-based neuro-fuzzy approach and swarm intelligence for
  time series forecasting,''
\newblock {\em Engineering Applications of Artificial Intelligence}, vol. 25,
  no. 2, pp. 295--308, 2012.

\bibitem{lins2013prediction}
Isis~Didier Lins, Moacyr Araujo, M{\'a}rcio das Chagas~Moura, Marcus~Andr{\'e}
  Silva, and Enrique~L{\'o}pez Droguett,
\newblock ``Prediction of sea surface temperature in the tropical atlantic by
  support vector machines,''
\newblock {\em Computational Statistics \& Data Analysis}, vol. 61, pp.
  187--198, 2013.

\bibitem{xie2019adaptive}
Jiang Xie, Jiyuan Zhang, Jie Yu, and Lingyu Xu,
\newblock ``An adaptive scale sea surface temperature predicting method based
  on deep learning with attention mechanism,''
\newblock {\em IEEE Geoscience and Remote Sensing Letters}, vol. 17, no. 5, pp.
  740--744, 2019.

\bibitem{kim2020sea}
Minkyu Kim, Hyun Yang, and Jonghwa Kim,
\newblock ``Sea surface temperature and high water temperature occurrence
  prediction using a long short-term memory model,''
\newblock {\em Remote Sensing}, vol. 12, no. 21, pp. 3654, 2020.

\bibitem{yuan2020deep}
Qiangqiang Yuan, Huanfeng Shen, Tongwen Li, Zhiwei Li, Shuwen Li, Yun Jiang,
  Hongzhang Xu, Weiwei Tan, Qianqian Yang, Jiwen Wang, et~al.,
\newblock ``Deep learning in environmental remote sensing: Achievements and
  challenges,''
\newblock {\em Remote Sensing of Environment}, vol. 241, pp. 111716, 2020.

\bibitem{sun2021time}
Yongjiao Sun, Xin Yao, Xin Bi, Xuechun Huang, Xiangguo Zhao, and Baiyou Qiao,
\newblock ``Time-series graph network for sea surface temperature prediction,''
\newblock {\em Big Data Research}, vol. 25, pp. 100237, 2021.

\bibitem{yu2020superposition}
Mei Yu, Zhuo Zhang, Xuewei Li, Jian Yu, Jie Gao, Zhiqiang Liu, Bo~You, Xiaoshan
  Zheng, and Ruiguo Yu,
\newblock ``Superposition graph neural network for offshore wind power
  prediction,''
\newblock {\em Future Generation Computer Systems}, vol. 113, pp. 145--157,
  2020.

\bibitem{qin2017dual}
Yao Qin, Dongjin Song, Haifeng Chen, Wei Cheng, Guofei Jiang, and Garrison
  Cottrell,
\newblock ``A dual-stage attention-based recurrent neural network for time
  series prediction,''
\newblock {\em arXiv preprint arXiv:1704.02971}, 2017.

\bibitem{wu2020connecting}
Zonghan Wu, Shirui Pan, Guodong Long, Jing Jiang, Xiaojun Chang, and Chengqi
  Zhang,
\newblock ``Connecting the dots: Multivariate time series forecasting with
  graph neural networks,''
\newblock in {\em KDD'20}, 2020, pp. 753--763.

\bibitem{LI2023119374}
ZhuoLin Li, Jie Yu, GaoWei Zhang, and LingYu Xu,
\newblock ``Dynamic spatio-temporal graph network with adaptive propagation
  mechanism for multivariate time series forecasting,''
\newblock {\em Expert Systems with Applications}, vol. 216, pp. 119374, 2023.

\bibitem{sutskever2014sequence}
Ilya Sutskever, Oriol Vinyals, and Quoc~V Le,
\newblock ``Sequence to sequence learning with neural networks,''
\newblock {\em Advances in neural information processing systems}, vol. 27,
  2014.

\bibitem{velivckovic2017graph}
Petar Veli{\v{c}}kovi{\'c}, Guillem Cucurull, Arantxa Casanova, Adriana Romero,
  Pietro Lio, and Yoshua Bengio,
\newblock ``Graph attention networks,''
\newblock {\em arXiv preprint arXiv:1710.10903}, 2017.

\bibitem{li2018diffusion}
Yaguang Li, Rose Yu, Cyrus Shahabi, and Yan Liu,
\newblock ``Diffusion convolutional recurrent neural network: Data-driven
  traffic forecasting,''
\newblock in {\em ICLR'18}, 2018.

\bibitem{pan2019urban}
Zheyi Pan, Yuxuan Liang, Weifeng Wang, Yong Yu, Yu~Zheng, and Junbo Zhang,
\newblock ``Urban traffic prediction from spatio-temporal data using deep meta
  learning,''
\newblock in {\em Proceedings of the 25th ACM SIGKDD international conference
  on knowledge discovery \& data mining}, 2019, pp. 1720--1730.

\bibitem{shang2021discrete}
Chao Shang, Jie Chen, and Jinbo Bi,
\newblock ``Discrete graph structure learning for forecasting multiple time
  series,''
\newblock {\em arXiv preprint arXiv:2101.06861}, 2021.

\bibitem{li2021dynamic}
Fuxian Li, Jie Feng, Huan Yan, Guangyin Jin, Depeng Jin, and Yong Li,
\newblock ``Dynamic graph convolutional recurrent network for traffic
  prediction: Benchmark and solution,'' 2021.

\bibitem{deng2021st}
Jinliang Deng, Xiusi Chen, Renhe Jiang, Xuan Song, and Ivor~W Tsang,
\newblock ``St-norm: Spatial and temporal normalization for multi-variate time
  series forecasting,''
\newblock in {\em Proceedings of the 27th ACM SIGKDD conference on knowledge
  discovery \& data mining}, 2021, pp. 269--278.

\end{thebibliography}

\end{document}